\documentclass[conference]{IEEEtran}
\IEEEoverridecommandlockouts

\usepackage{cite}
\usepackage{amsmath,amssymb,amsfonts}
\usepackage{algpseudocode}
\usepackage{textcomp}
\usepackage{xcolor}
\usepackage{soul}
\usepackage{graphicx} 
\usepackage{comment}
\usepackage{pdfpages}
\usepackage{hyperref}

\usepackage[left=1.62cm,right=1.62cm,top=.75in]{geometry}

\def\BibTeX{{\rm B\kern-.05em{\sc i\kern-.025em b}\kern-.08em
    T\kern-.1667em\lower.7ex\hbox{E}\kern-.125emX}}

\title{Generating Synthetic Time Series Data for Cyber-Physical Systems\\
\thanks{This material is based upon work supported by the Engineering Research and Development Center - Information Technology Laboratory (ERDC-ITL) under \#W912HZ23C0013. Any opinions, findings and conclusions or recommendations expressed in this material are those of the author(s) and do not necessarily reflect the views of the ERDC-ITL.}
}

\makeatletter
\newcommand{\linebreakand}{%
  \end{@IEEEauthorhalign}
  \hfill\mbox{}\par
  \mbox{}\hfill\begin{@IEEEauthorhalign}
}
\makeatother

\newbool{Long_names}

\ifbool{Long_names}
{
\author{
\IEEEauthorblockN{Alexander Sommers}
\IEEEauthorblockA{\textit{Computer Science and Engineering} \\
\textit{Mississippi State University}\\
ams1988@msstate.edu}
\and
\IEEEauthorblockN{Somayeh Bakhtiari Ramezani}
\IEEEauthorblockA{\textit{Computer Science and Engineering} \\
\textit{Mississippi State University}\\
sb3182@msstate.edu}
\and
\IEEEauthorblockN{Logan Cummins}
\IEEEauthorblockA{\textit{Computer Science and Engineering} \\
\textit{Mississippi State University}\\
nlc123@msstate.edu}
\linebreakand
\IEEEauthorblockN{Sudip Mittal}
\IEEEauthorblockA{\textit{Computer Science and Engineering} \\
\textit{Mississippi State University}\\
mittal@cse.msstate.edu}
\and
\IEEEauthorblockN{Shahram Rahimi}
\IEEEauthorblockA{\textit{Computer Science and Engineering} \\
\textit{Mississippi State University}\\
rahimi@cse.msstate.edu}
\linebreakand
\IEEEauthorblockN{Maria Seale}
\IEEEauthorblockA{\textit{Eng. Research and Dev. Center} \\
\emph{Department of Defence} \\
maria.a.seale@erdc.dren.mil}
\and
\IEEEauthorblockN{Joseph Jaboure}
\IEEEauthorblockA{\textit{Eng. Research and Dev. Center} \\
\emph{Department of Defence} \\
joseph.e.jabour@erdc.dren.mil}
}
}
{
\author{
\IEEEauthorblockN{Alexander Sommers}
\IEEEauthorblockA{\textit{Comp. Sci. \& Eng.} \\
\textit{Mississippi State University}\\
ams1988@msstate.edu}
\and
\IEEEauthorblockN{Somayeh Bakhtiari Ramezani}
\IEEEauthorblockA{\textit{Comp. Sci. \& Eng.} \\
\textit{Mississippi State University}\\
sb3182@msstate.edu}
\and
\IEEEauthorblockN{Logan Cummins}
\IEEEauthorblockA{\textit{Comp. Sci. \& Eng.} \\
\textit{Mississippi State University}\\
nlc123@msstate.edu}
\linebreakand
\IEEEauthorblockN{Sudip Mittal}
\IEEEauthorblockA{\textit{Comp. Sci. \& Eng.} \\
\textit{Mississippi State University}\\
mittal@cse.msstate.edu}
\and
\IEEEauthorblockN{Shahram Rahimi}
\IEEEauthorblockA{\textit{Comp. Sci. \& Eng.} \\
\textit{Mississippi State University}\\
rahimi@cse.msstate.edu}
\and
\IEEEauthorblockN{Maria Seale}
\IEEEauthorblockA{\textit{Eng. Research and Dev. Center} \\
\emph{Department of Defence} \\
maria.a.seale@erdc.dren.mil}
\and
\IEEEauthorblockN{Joseph Jaboure}
\IEEEauthorblockA{\textit{Eng. Research and Dev. Center} \\
\emph{Department of Defence} \\
joseph.e.jabour@erdc.dren.mil}
}
}

\begin{document}

\maketitle
\begin{abstract}
Data augmentation is an important facilitator of deep learning applications in the time series domain. A gap is identified in the literature, demonstrating sparse exploration of the transformer, the dominant sequence model, for data augmentation in time series. A architecture hybridizing several successful priors is put forth and tested using a powerful time domain similarity metric. Results suggest the challenge of this domain, and several valuable directions for future work.
\end{abstract}


\section{Introduction}
Deep neural networks can model the complex functions implicit in time series data, but require high volumes of data, which may not always be available. Data augmentation can be used to supplement relatively small datasets, and can employ powerful sequence models, such as the transformer neural network (TNN) \cite{transformer_origin} . 

This work's contributions are:
\begin{itemize}
    \item A survey of the intersection of time series generation and the transformer archatecture, identifying a gap.
    \item A framework for a pure transformer time series transformer synthisizer to achieve data augmentation combining powerful mechanisms from successful prior models.
    \item A novel derivation of the Wasserstein Fourier Distance for evaluating synthesis quality based on frequency domain features.
\end{itemize}


\section{Background}

Date augmentation (DA) is a countermeasure to data-deprivation caused by confidentiality concerns, or prohibitive collection expense. DA methods can be divided into \emph{permutation} and \emph{generation} methods. Permutation methods edit extant data to make more (e.g. jittering, time-warping)\cite{time_series_data_aug_survey}. Generative methods have proven more successful, and include generative adversarial networks (GANs). 

Successful GANs for time series DA exist using CNN and RNN backbones \cite{time_series_data_aug_survey, TimeGAN}. But the TNN has become the dominant sequence model recently.  The attention mechanism allows for long range pairwise context modeling, needing neither convolution or a hidden state. They are so successful that TNN based large language models (LLMs) are nearly synonymous with the state of the art in machine learning, at time of writing. This motivates the present work.

\section{Prior Art and Motivation}
Time series DA using pure transformer GANs seems to be relatively unexplored compared to DA in the image domain, and time series DA using RNN backbones \cite{TimeGAN}. The works of Li et al. \cite{Close_competitor, Closest_competitor} are an exception, and use inferences from \cite{TransGAN} and \cite{Patching} to augment their potential. 

The novelty of transformer based time series DA was established by reviewing several comprehensive surveys. Iglesias et al. \cite{time_series_data_aug_survey} describes the use of GANs in this domain, but with no category for transformer based GANs. Esteban et al. \cite{Medical_cGANs} surveyed the use of DA in the health domain. Unlike \cite{time_series_data_aug_survey}, they looked beyond time series data, but no presented GANs are transformer based. Figueira et al. \cite{Figueira2022SurveyOS} present most GANs as in the image domain. The domain of tabular data is mentioned, and a minority of the GANs presented can perform on it. Specific transformations were needed to apply an LSTM based tabular GAN to its data \cite{TGAN}. The necessity of this step supports the apparent distance between GANs and time series data. Wen et al. \cite{tranf_in_TS_tut} does not reference such preparations, though its attention to tabular data is  low, and it is worth noting that time series data is often thought of as distinct from tabular data. These surveys demonstrate sparse exploration of the intersection of transformers, GANs, and time series DA.

Both \cite{time_series_data_aug_survey} and \cite{Closest_competitor} report the application of image domain methodologies into time series applications. The domains of image and sequence generation have been explored with transformer based GANs. This provides precedent and opportunity for the application of already established methods from these domains to the task of time series DA. Forecasting in sequences and time series, such as in the work of Zhou et al. \cite{Informer} also demonstrate the successful applicability of transformers to long time series data.

This apparent sparseness of exploration, and the abundance of potential inferences into the generative task from these related domains, motivate the present work.

\section{Methodology}\label{Methodology}

The present architecture is a pure transformer \cite{transformer_origin} following the examples of \cite{Closest_competitor, Close_competitor}, \cite{Informer}, \cite{TransGAN}, and \cite{ACGAN_for_MFD}. The architecture was constructed in anticipation of certain features of the FEMTO dataset \cite{PRONOSTIA}, used in the primary experiments. It is a prognostic machine health dataset describing the run to failure of ball bearings monitored by accelerometers. Section \ref{cgen} discuses the mechanisms of conditional generation, and section \ref {imp} certain implementation details. Then, section \ref{WFD} discusses the method selected to evaluate the output, and section \ref{exp} presents details of the experiments. 

\subsection{Model Architecture}\label{architecture}

\subsubsection{Conditioned Generation}\label{cgen}

\begin{figure}[h]
\centering
\includegraphics[width=0.4\textwidth]{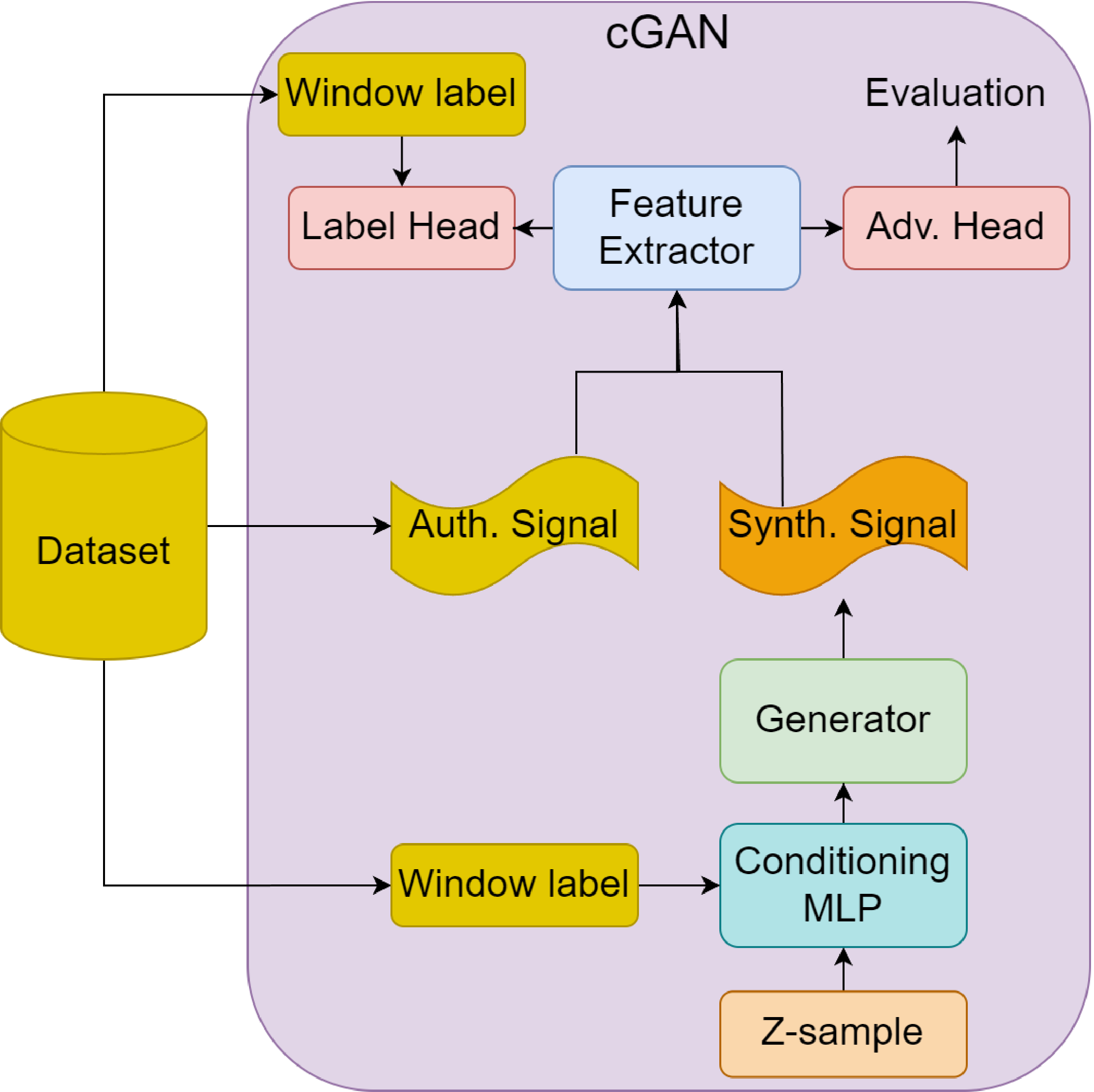}
\caption{Each data point consists of a window (authentic signal) over time, and a label. The generator accepts a noise sample conditioned on the label, and the discriminator outputs evaluations of ``realness" \cite{WGAN_origin}, and a label to be compared to the target conditional.}
\label{Gross}
\end{figure}

Each step can be followed in figure \ref{Gross}. GANs  have two primary components, the ``generator" and the ``discriminator" \cite{GAN_origin}. The generator must learn the feature dependencies within the target dataset, then being used to synthesize alternatives to the authentic data. The loss needed to train the generator is supplied indirectly using the discriminator. The discriminator is directly trained to distinguish between synthetic and authentic samples. The present task requires greater specificity in its generation, adopting the conditional-GAN (cGAN) \cite{cGAN_origin} paradigm. 

\begin{figure}[h]
\centering
\includegraphics[width=0.45\textwidth]{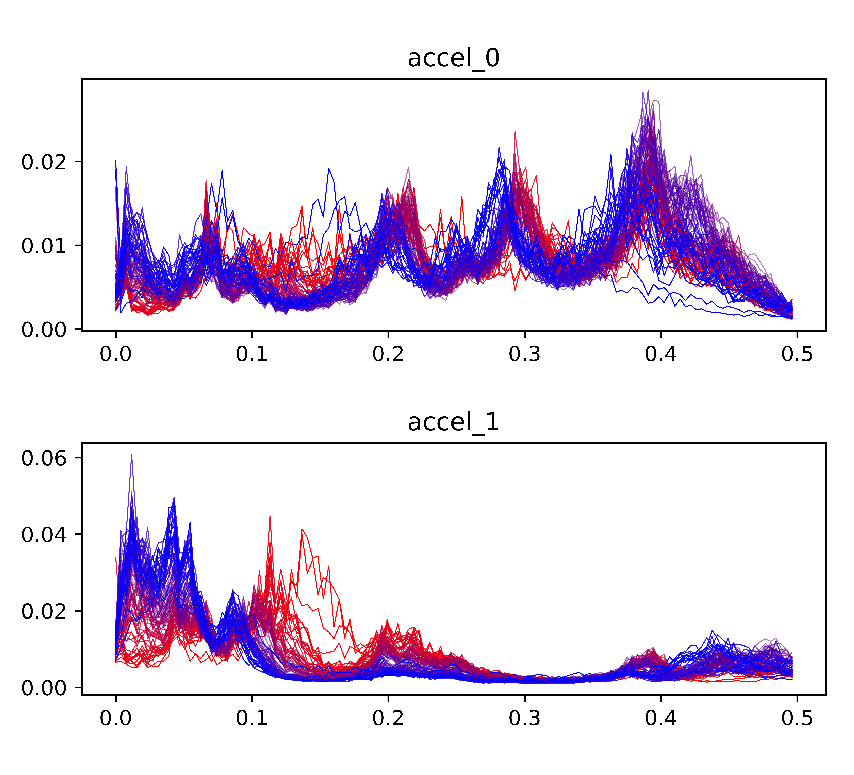}
\caption{The normalized power spectrums \cite{WFD} for the accelerometers of bearing 1 run 1 of the FEMTO dataset, taken at evenly spaced windows across the run. Blue indicates the beginning, red the end. While the shifts upwards in frequency are subtle in the first accelerometer, they are clear in the second. The depicted are not strict correspondents to actually observed frequencies, and so the lack of units on the x axis is intentional.}
\label{freq_shift}
\end{figure}

cGANs require conditional inputs to influence the stochastic generative process. This conditional information is commonly a class label. If a target time series lacks labeled segments, then an alternative must be employed. In the present case, distinct shifts in frequency over time were observed, as shown in figure \ref{freq_shift}. These shifts are a function of the bearing, the running conditions of the experiment, and the degradation of the bearing over time. Thus, the label supplied this information to the generator. A one hot encoding was used to encode bearing and condition information, since there were a small number of experiments \cite{PRONOSTIA}. An appended real value encoded the percent of bearing lifetime which had passed by the start of the labeled window. Each window in the data set for all bearings was paired with its own conditional label. All training was conducted using the Pytorch Lightning Library \cite{Falcon_PyTorch_Lightning_2019} using a NVIDIA G-FORCE 1080Ti GPU. The conditioning scheme is not the canonical form, instead following the asymmetrical scheme used by Li et al. \cite{Closest_competitor}.

The conditioning MLP prior to the generator accepts the noise sample and the conditioning label. The noise vector is of size 100, and the label is projected up to a size of 500 before the two are concatenated and projected to the appropriate size for the generator \cite{cGAN_origin}. The generator outputs tensors of the same length and channel depth as the target windows. All windows have the same number of time steps, which is always a power of two to be compatible with the up-scaling scheme the generator uses. The synthetic windows denormalized during inference for use elsewhere. During training they are instead passed to the ``critic". 

The Wasserstine GAN loss configuration \cite{WGAN_origin} uses a ``critic" instead of a ``discriminator" because the output of the adversarial head of the discriminator is not a probability that some input window is from the authentic dataset, but rather a score of how ``real" it seems. The critic contains a feature extraction pipeline based on the encoder from Zhou et al.'s Informer model \cite{Informer}, and terminates in two output heads. The ``adversarial head" outputs the ``realness" score of the input window, while the ``label" head outputs an estimate of the label corresponding to the input window. The losses for training the critic and generator are themselves adversarial, producing competition between the networks.

In Li et al.'s loss implimentation, the gradient penalty WGAN loss \cite{WGAN_2} is modified to combine the loss for the two heads. The critic is penalized for granting high realness scores to synthetic windows, relative to the scores given to authentic windows, and for mislabeling the authentic windows with its label head. The weighted MSE loss is used to penalize the deviance between the expected and predicted label. 

The generator's objective is to produce synthetic samples that cause the adversarial head of the critic to output a high realness score, and cause the label head to return the same label with which the generation was conditioned. The critic is penalized when its outputs receive low realness scores and labels unlike those used to condition them.

\subsubsection{Implementation Details}\label{imp}

\begin{figure}[h]
\centering
\includegraphics[width=0.4\textwidth]{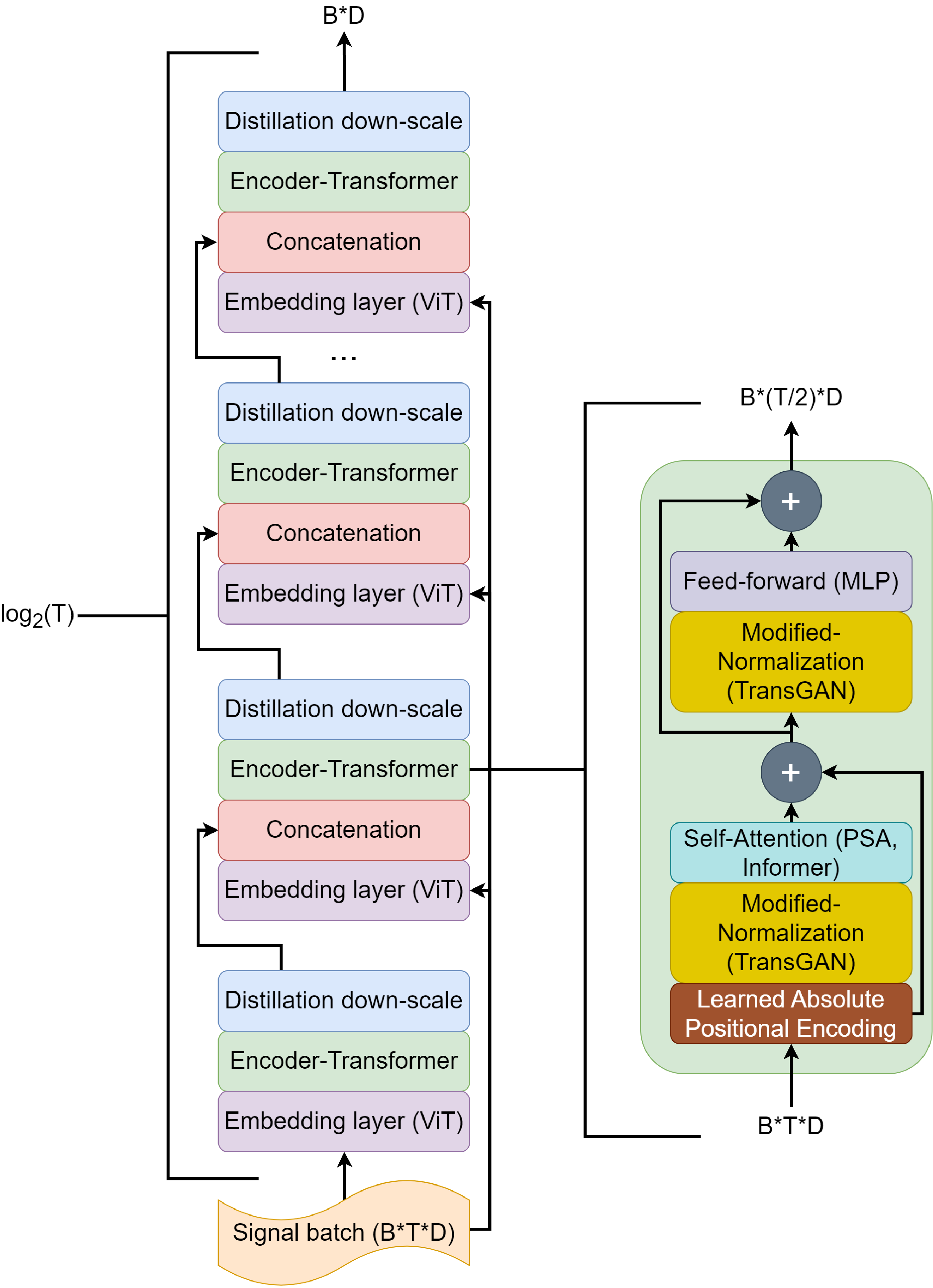}
\caption{While T, the number of time steps, is a power of 2, it takes $\text{log}_{2}(\text{T})$ blocks in the extraction pipeline to reduce each window to a tensor of one less rank. B is batch size and D channel depth. On the right is a break out of the employed transformer module.}
\label{Extraction}
\end{figure}

The feature extraction pipeline forming the bulk of the critic is illustrated in figure \ref{Extraction}. From the discriminator of Jiang et al.'s TransGAN \cite{TransGAN}, the feature extractor adopts a hierarchical series of transformers designed to extract features from the data at coarsening levels of granularity. This is achieved using the patching method introduced by Dosovitskiy et al.'s ViT \cite{Patching}. Transformers earlier in the hierarchy attend to detailed features (small patch size) and concatenate their output to the input of the next transformer (larger patch size). Each transformer sub-module uses Informer's probabilistic sparse attention (PSA) and distillation to avoid the quadratic cost of canonical attention on long time-series data. All transformer blocks use the instance norm proposed in \cite{TransGAN} and the pre-norm residual arrangement suggested by Xoing et al. \cite{Pre_LN}.

\begin{figure}[h]
\centering
\includegraphics[width=0.15\textwidth]{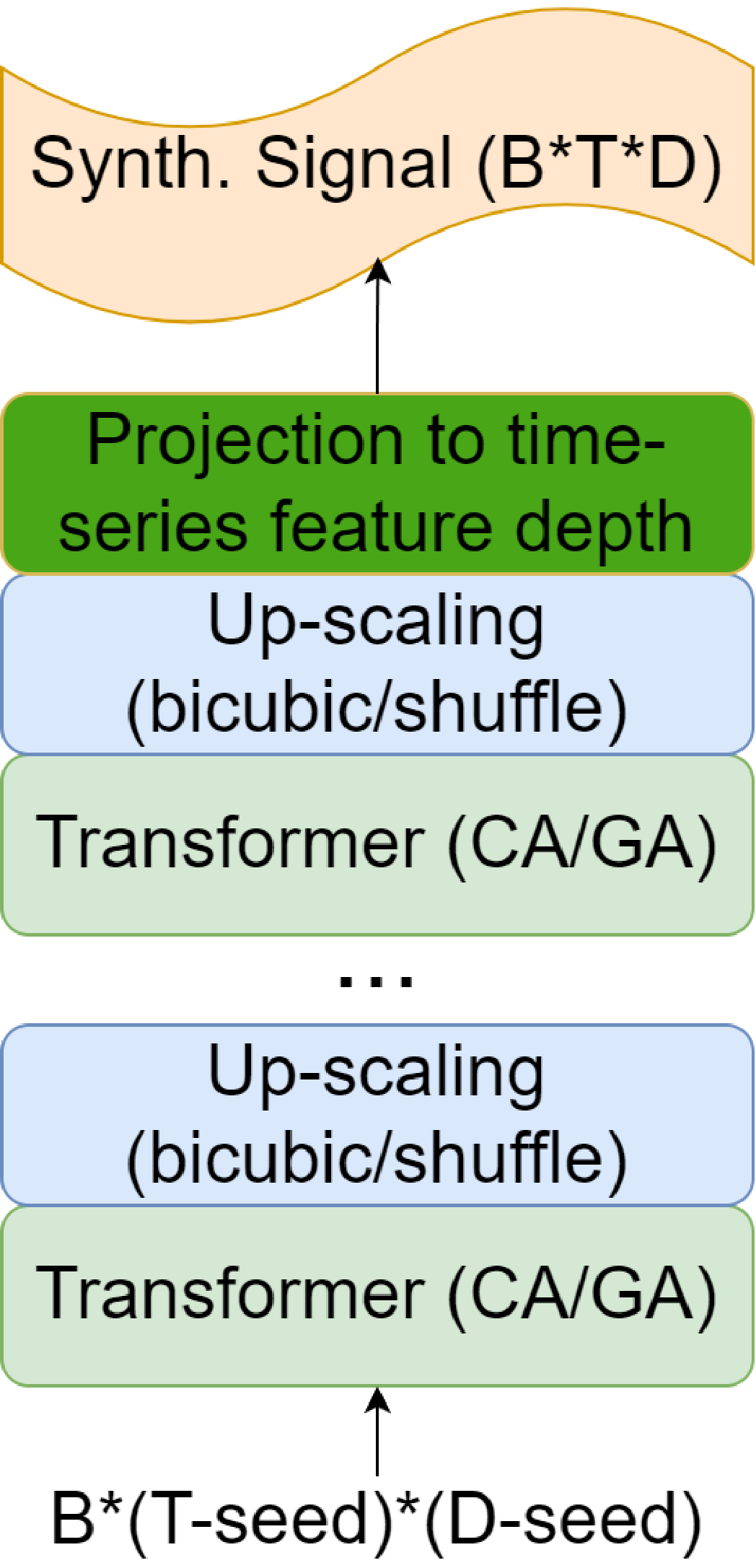}
\caption{The seed dimensions (T-seed, D-seed) are set using hyper parameters. A smaller seed allows for more layers of up-scaling, allowing for greater learning capacity as the number of learned parameters increases. This also increases the computational cost of generation.}
\label{GEN}
\end{figure}

The generator, described in figure \ref{GEN}, is closely derived from TransGAN, with two salient differences. First, there are no reshaping operations needed to move back and forth between an ``image form" (TransGAN operates in the image domain) and patched-sequence form of the data. Secondly, though grid attention (GA) is employed, it is not achieved using distinct, parallel transformers. Having a separate transformer for each grid partition is a powerful inductive bias when grid positions are likely to be non-trivially associated with certain features of the data. This is clearly true in, for example, images of faces. There, the face tends to be centered, looking at the camera. Thus, grid positions in the center will concern facial features, while peripheral ones will not. In the present case, time series segments did not appear to present an opportunity to use this bias, which also vastly increases model size. As such, grid attention is achieved by splitting inputs windows into grid partitions, performing attention within each partition, and then reassembling the windows. 

GA was employed over PSA because there was the possibility that PSA would homogenize the more trivial details of the output. Details can be trivial, but they are still part of a complete synthetic output. \cite{TransGAN} shows that GA is capable of preserving details, despite its local bias. There are two threshold values at play in the generator. The generator iterativley up-scales from T-seed and D-seed to the target time and channel size respectively. The can save a great deal of time and memory cost and was used in the image domain to generate large, high resolution imagery. The shuffle-threshold determines at which size of T the shuffle operation \cite{pix_shuffle} begins to be employed, instead of bicubic interpolation. Bicubic interpolation doubles the time dimension of the signal. The shuffle operation does this as well, but simultaneously halves the feature depth. For this reason, D-seed is often quite large (1024 in \cite{TransGAN}). The second threshold is the grid attention-threshold. This determines at which size of T the network begins to employ GA to avoid quadratic attention costs. Canonical attention is used below the grid threshold.

Both the critic and generator employ rotary position embedding (RoPE), as introduced by Su et al. \cite{RoPE}, and soft/learned absolute positional embedding (LaPE) \cite{TransGAN, Closest_competitor}. The former are used when computing the affinity matrix of each self-attention operation, while the latter are applied after scaling and concatenation operations, or when a time-series tensor is first introduced into a pipeline. The training process in \cite{TransGAN} employed a large number of transformers trained on quite powerful hardware. Due to hardware, time, and dataset constraints, the present work follows the more minimal lead of \cite{Informer}, with only one transformer block between scaling operations. All transformers in the present model are strictly encoder style transformers, which possess self-attention and feed-forward components, but lack encoder-decoder attention. 

The both critic heads are MLPs employing the leaky rectified linear unit (LeReLU) activation function between layers. The label head terminates in a sigmoid activation function, while the adversarial head has no terminal activation function \cite{WGAN_origin}. Each head has a single hidden layer, loosely modeled on the head in \cite{Patching}. The one-hot portion of the label vector employs label smoothing \cite{smoothing}. By the time data exits the feature extraction pipeline, it has been reduced from a rank three to a rank two tensor, collapsing the temporal rank through repeated halving.

\subsection{Output Evaluation}\label{WFD}

Once a GAN has been trained, its output must be evaluated to determine its suitability for DA. Methods of evaluating GAN output have been concentrated in the visual domain, such as the well known inception score \cite{Inception}. A derivative of Cazelles et al.'s Wasserstein-Fourier distance (WFD) metric was used in the present work \cite{WFD}. It was selected based on the clear frequency shifts seen concurrent to bearing degradation. Details are given in the origin work, but an overview of the derivation is provided below. The NumPy library \cite{numpy} was used to construct the metric derivation in the present work.

Signals over time can subjected to the Fourier transform. That transform can be rendered into a power spectrum known as the power spectral density (PSD). The area under the power vs. frequency curve is the total ``power". Dividing all powers by this sum normalizes the PSD into the normalized PSD (NPSD). Under conditions of extant and finite power, an NPSD can be treated as a probability distribution since the sum of its powers is one \cite{WFD}. The Wasserstein distance (``earth movers" distance) can be computed for any two probability distributions of the same dimensionality. The most basic WFD is between two one-dimensional signals over time, each converted into NPSDs. This distance is in the range $[0,\infty)$, and forms the foundation of the present output evaluation. The following sections describe derivations of the WFD the present work uses.

\subsubsection{Distances Between Segments}
A segment is comprised of $D$ distinct features over $T$ contiguous time steps. Two time series segments with one feature each are defined as more similar if the NPSDs corresponding to their respective signals have a low WFD. If the segments have more than one feature, this operation is performed across features and the arithmetic average is taken. Let the WFD between two sequences $A$ and $B$, or their NPSDs, will be $W(A,B)$, with the understanding that this is the average distance between the corresponding features of $A$ and $B$. It should be noted that this derivation is not an extension of the WFD into multiple dimensions as the Wasserstein distance (the Wasserstein-2 distance in this case) is not computed between two multi-dimensional NPSDs but between several mono-dimensional NPSDs. This suffices for pairwise segment similarity measures, but more is required to characterize collections of segments.

\subsubsection{The Mean NPSD and the Standard Distance}
A set of $n$ windows can be converted into NPSDs, and these can then be averaged. If a given NPSD $s_{x}$ is the $x$\textsuperscript{th} NPSD of a set $K$, then $\overline{s_K} = \frac{\sum_{i = 1}^{n}s_i}{n}$ is the mean NPSD. Coherent sets of segments include samples from neighborhoods around a target, and sets of synthetic outputs on a single target. Once the mean has been computed, the standard distance of $K$ can be computed as $\sigma = \frac{\sum_{i = 1}^{n} {W(s_{i}, \bar{s_{K}})}}{n}$. $\sigma$ allows for the characterization of how diverse $K$ is, and is analogous to the standard deviation of a set of numbers. Larger $\sigma$ suggest that $\overline{s_K}$ less accurately characterizes the population. The standard distance provides an intraset measure of variance, and the WFD between the mean NPSDs of two sets provides a measure of interset variance. This suffices for quantitative measures, but also supplies qualitative data since NPSDs, as power spectrums, are common tools for visual analysis of data.

\subsection{Experiments}\label{exp}

\begin{figure}[h]
\centering
\includegraphics[width=0.45\textwidth, height=.7\textwidth]{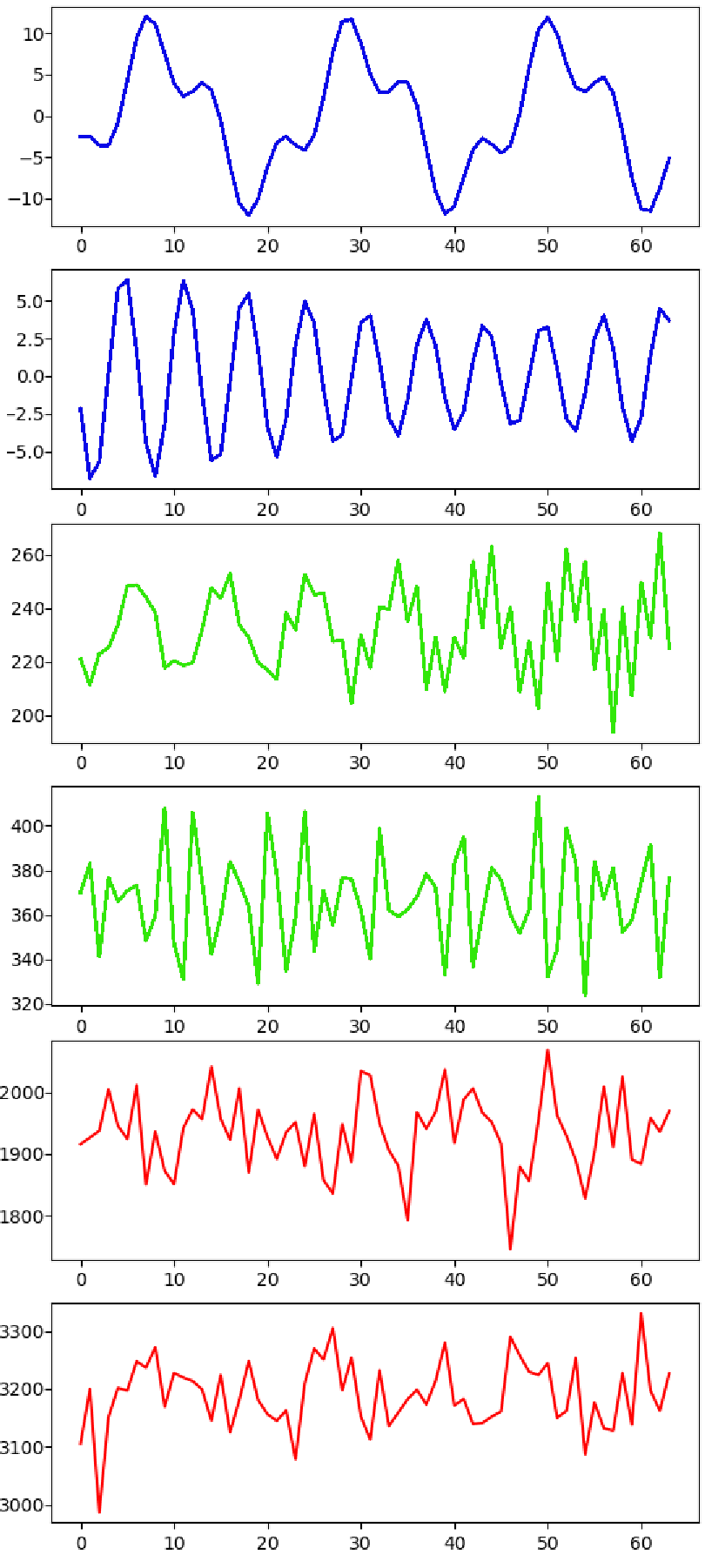}
\caption{The blue, green, and red windows are instances of the easy, medium, and hard artificial datasets. Each window has to temporally concurrent features, similar to the two accelerometer features in the FEMTO dataset. All windows are produced using compounded sine waves, and each class had a non-overlapping range of features used to stocastically generate each instance.}
\label{art_dat}
\end{figure}

Primary experiments were conducted using the vertical and horizontal accelerometer data from the FEMTO dataset \cite{PRONOSTIA}. All data were subdivided into windows of 256 contiguous time steps, yeilding a total of 248,890 datapoints. These were divided into a 20\%, 70\%, 10\% test-train-validate split. To avoid data snooping, each subset had its own normalization parameters. 

A secondary, artificial, dataset was created using compounded sine-waves. An evenly balanced dataset of 50,000 artificial compound waves, 64 time steps long each, was generated. There were three levels of compounding reflecting three different levels of challenge for the GAN to model. Though generation was stocastic, the three classes had distinct frequency, amplitude, phase shift, and compounding (number of compounded waves) ranges (examples seen in figure \ref{art_dat}). Conditioning labels for the artificial dataset were simple three element one hot encodings of the class of the window (easy, medium, hard). Experiments using the artificial data used the same training split as those using the FEMTO data.

\section{Results}

\begin{figure}[h]
\centering
\includegraphics[width=0.45\textwidth]{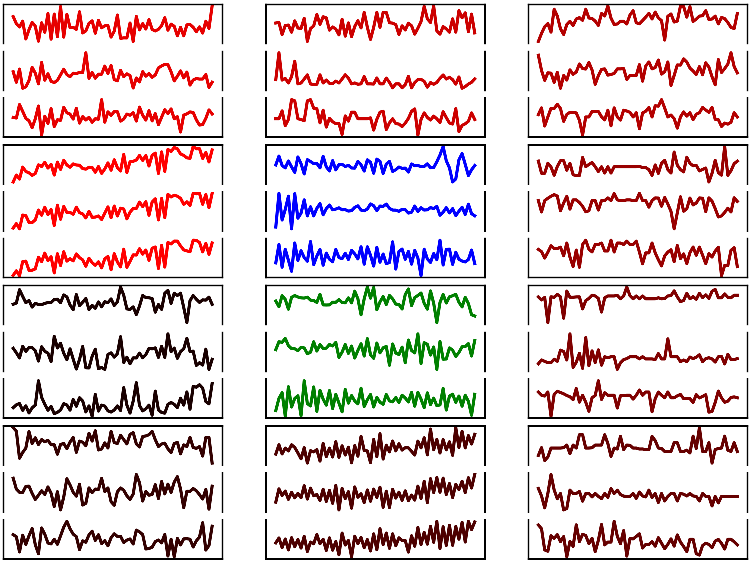}
\caption{ In each cell are three windows over a single accelerometer sensor. Blue windows were collected from a Gaussian-distribution around a target point in the bearing data to explore variation in generative behavior around a target point. Each blue window was then a target for generation synthetic counterpart. A green window neighbors its corresponding blue window, to give an indication of variability around each selected target. Moving clockwise from red to black across training epochs, the outputs transition from homogeneous to visually plausible synthetic. Only three targets are shown to facilitate ease of visualization.}
\label{only_random}
\end{figure}

Despite using promising mechanisms, and thorough checking of original and secondary codebases, the proposed architecture performed poorly on the FEMTO dataset. This is also surprising since prior works applying GANs to similar data performed acceptably \cite{GPDH1, GPDH2, GPDH3, GPDH4}. Initial results seemed promising as seen in figure \ref{only_random}. However, using the proposed evaluation metric, it was found that the synthetic windows only appeared to match the visual character of the target outputs. This illustrates the peril of attempting visual assessment of this kind of data. Authentic windows appeared quite random, but are not. While synthetic data did transition from implausibly uniform to visually plausible outputs, inspection with frequency decomposition revealed that the generator had only learned to produce random jagged data, lacking relevant frequency domain correspondence to the target dataset. This does affirm the use of frequency domain features in this domain.

Given that generation is nearly always more challenging than classification, it is unsurprising that the FEMTO dataset would prove difficult to mimic. Tests were also run on the artificial dataset, but results were also poor, with high averaged Wasserstein Fourier Distances. 

\section{Discussion and Future Work}

It is surprising that elsewhere successful components failed to perform in this application. The planned future work will focus on a diagnostic evaluation of the poor performance of the constituent mechanisms. Surprisingly high performance is always desired, but surprisingly poor performance, well explained, is usually more edifying \cite{TEIH}.

Particularly, the compatibility of approximation methods meant to favor local/sparse attention with something as laden with long range dependencies as time series synthesis should be investigated. Countermeasures to the quadratic cost of canonical attention might actually undermine the arbitrary distance modeling attention can achieve at long-range. The poor results could also be due to low learning capacity. The employed hardware, inadequate to train now commonly large numbers of hyper parameters, will need to be upgraded for exploration of this hypothesis. Lastly, it may be that the entire notion of a pure transformer for this task needs to be revisited in light of diffusion models, and considered alongside the apparent non-convergence of this project's core prior architecture in other tests \cite{TimeWeaver}. 

\bibliographystyle{ieeetr} 
\bibliography{ref} 

\end{document}